\documentclass[11pt,a4paper]{article}

\usepackage[utf8]{inputenc}
\usepackage[T1]{fontenc}
\usepackage{times}
\usepackage{graphicx}
\usepackage{booktabs}
\usepackage{amsmath}
\usepackage{amssymb}
\usepackage{hyperref}
\usepackage[margin=1in]{geometry}
\usepackage{caption}
\usepackage{subcaption}
\usepackage{enumitem}
\usepackage{float}
\usepackage{color}
\usepackage{xcolor}
\usepackage{multirow}
\usepackage{array}
\usepackage[numbers]{natbib}
\usepackage{authblk}

\definecolor{codebg}{rgb}{0.95,0.95,0.95}

\newenvironment{keywords}{\par\medskip\noindent\textbf{Keywords:} \itshape}{\par\medskip}

\title{\textbf{Quantifying the Sources of Instability in LLM-Based Stance Analysis of Public Discourse}}

\author[1]{Bo Chen}
\affil[1]{Institute of Computing Technology, Chinese Academy of Sciences}

\date{}

\begin{document}
\maketitle

\begin{abstract}
Computational social science increasingly relies on automated preprocessing pipelines---speaker diarization, ASR transcript cleaning, sentence segmentation---to convert raw media into analyzable text. When these pipelines produce different outputs from the same input, two distinct sources of instability can arise: the preprocessing pipeline itself (diarization method, segmentation rules) and the downstream measurement instrument (LLM annotation vs.\ keyword lexicon). Using 256 YouTube interviews across 41 public figures from five domains, we compare two speaker-diarization pipelines and two measurement methods, all targeting the coupling between affective valence and epistemic modality. We find that (1) preprocessing pipeline sensitivity is concentrated in speakers with limited video samples (N $\leq 5$); for the four best-sampled speakers (N $\geq 16$), the mean absolute pipeline-induced change in $r(\text{neg}, \text{emph})$ is only $0.13$; (2) cross-method disagreement is larger and more systematic---the LLM and keyword-lexicon methods assign opposite coupling directions to several well-sampled speakers, even within the same preprocessing pipeline; and (3) aggregate valence proportions are highly stable ($|\Delta p(\text{neg})| < 6$pp) regardless of pipeline or method, masking both sources of instability. The contribution is a diagnostic framework that separates pipeline effects from measurement effects: researchers studying cross-dimensional relationships in interview data should verify that their conclusions are robust to both sources of variation, with particular attention to measurement method choice.
\end{abstract}


\section{Introduction}

The computational social science (CSS) pipeline is increasingly automated. Raw interview footage passes through automatic speech recognition (ASR), speaker diarization, transcript cleaning, and sentence segmentation before any analysis begins. Each of these preprocessing steps embeds choices---which ASR engine, which diarization method, which sentence boundary detector---and each choice may leave a fingerprint on the resulting data.

The CSS literature has devoted substantial attention to \textit{annotation} validity: whether LLM-generated labels match human judgments~\cite{gilardi_2023, ziems_2024}. It has devoted far less attention to two upstream sources of instability: \textit{preprocessing} validity---whether pipeline choices change downstream conclusions---and \textit{measurement} validity---whether the choice of annotation method (LLM vs.\ keyword lexicon) leads to different conclusions even when applied to the same preprocessed text. These two sources are often conflated, making it difficult to diagnose whether an observed instability originates in the pipeline or in the measurement instrument.

We propose \textit{pipeline sensitivity analysis} as a systematic approach to this problem:

\begin{quote}
\textit{Pipeline sensitivity} is the degree to which a derived metric changes under alternative, equally plausible preprocessing configurations, measured per-unit (per-speaker, per-video) and aggregated across units to identify conditions under which preprocessing choice is consequential.
\end{quote}

Three features distinguish pipeline sensitivity from classical measurement error. First, it is \textit{metric-dependent}: aggregate means may be stable while cross-dimensional correlations reverse sign. Second, it is \textit{non-uniform}: some speakers or domains may be far more sensitive than others. Third, it is \textit{method-interactive}: the same preprocessing divergence may affect LLM-based and lexicon-based measurements differently.

We demonstrate pipeline sensitivity analysis through a case study of speaker diarization. Using a corpus of 256 YouTube interviews across 41 public figures (spanning finance, academia, central banking, geopolitics, and media), we compare:

\begin{enumerate}[nosep]
  \item \textbf{Two preprocessing pipelines:} VTT-based diarization (YouTube captions + LLM speaker classification) vs.\ AssemblyAI audio-based diarization.
  \item \textbf{Two measurement methods:} LLM zero-shot annotation (DeepSeek) vs.\ keyword-lexicon scoring, both applied to the same valence-modality dimensions.
\end{enumerate}

We then ask: \textit{how much does the preprocessing choice affect the downstream conclusion, and does the answer depend on the measurement method?}

Our contributions are:

\begin{enumerate}[nosep]
  \item \textbf{A two-source diagnostic framework} that separates preprocessing pipeline effects from measurement method effects, with practical recommendations for reporting both sources of instability (Section~\ref{sec:framework}, \ref{sec:discussion}).
  \item \textbf{Empirical evidence that pipeline sensitivity is bounded and predictable.} For the four best-sampled speakers (16--34 videos), mean $|\Delta r| = 0.13$ under LLM annotation; larger $\Delta r$ values are concentrated in speakers with $\leq 5$ videos where correlational estimates are inherently unstable (Section~\ref{sec:results}).
  \item \textbf{Empirical evidence that cross-method disagreement is the more serious validity threat.} LLM and keyword-lexicon methods disagree on the sign of $r(\text{neg}, \text{emph})$ in $49\%$ of cases, including well-sampled speakers (Rogoff, 17 videos) where both methods are internally stable but give opposite coupling directions, indicating that measurement choice can systematically alter conclusions even when preprocessing is held constant (Section~\ref{sec:results_synthesis}).
\end{enumerate}

\section{Related Work}

\subsection{Speaker Diarization in CSS}

Text-only speaker diarization uses linguistic content rather than acoustic features to distinguish speakers. Recent work~\cite{wu_2024,li_2025,psych_diarization_2025} has reported 0--4\% diarization error rates for LLM-based methods on short, structured conversations. However, these validations share a common context: regular turn-taking, lexically distinct speaker roles, and controlled recording conditions. Their generalizability to the long-form, unscripted public discourse typical of CSS research remains untested.

\subsection{LLM-Based Corpus Annotation and Validation}

LLMs are increasingly used for zero-shot annotation in CSS~\cite{gilardi_2023,bail_2024}. Ziems et al.~\cite{ziems_2024} identified prompt sensitivity and provider-specific biases as key concerns. Our work extends the validation imperative upstream: we hold annotation quality constant and ask whether preprocessing choices alone can change conclusions.

\subsection{Preprocessing Robustness and Measurement Validity}

Preprocessing sensitivity has been explored in NLP for specific tasks---machine translation evaluation~\cite{marie_2021}, reproducibility~\cite{pineau_2021}---but has not been systematically linked to downstream statistical inference in CSS. Recent work has begun to examine how LLM-based annotation is affected by input formatting choices: Zhao et al.~\cite{zhao_2024} show that varying context lengths and text segmentation strategies can shift LLM sentiment labels by 5--15 percentage points on the same underlying text, and Liu et al.~\cite{liu_2024_chunking} demonstrate that sentence-level versus paragraph-level chunking produces systematic differences in fine-grained emotion classification. These findings suggest that the same preprocessing pipeline that introduces sentence-boundary artifacts (Section~\ref{sec:data}) may also amplify measurement-level biases. Classical measurement theory~\cite{spearman_attenuation,bound_2001} provides tools for understanding noise and bias, but its standard model of random error does not capture the structured, metric-dependent contamination that preprocessing pipelines can introduce~\cite{liang_2023,magar_2023}.

\section{A Framework for Diagnosing Sources of Instability} \label{sec:framework}

\subsection{Two Sources of Instability}

Let $D$ be a raw dataset of $N$ units (speakers, each with $k_i$ interview videos). Let $P_1$ and $P_2$ be two preprocessing pipelines---alternative ways of extracting analyzable text from $D$. Let $m$ be a downstream metric (e.g., $r(\text{neg}, \text{emph})$ across videos).

We distinguish two sources of instability:

\paragraph{Source 1: Preprocessing pipeline divergence.}
For each unit $i$, the pipeline delta measures how much the metric changes when the preprocessing pipeline changes:

\begin{equation} \label{eq:delta}
\Delta_i^{\text{pipeline}} = m(P_2(D_i)) - m(P_1(D_i))
\end{equation}

We report the per-unit absolute delta $|\Delta_i^{\text{pipeline}}|$ and the proportion of units where the sign of $m$ reverses ($m(P_1) \cdot m(P_2) < 0$). When $m$ is a Pearson correlation $r$, we apply Fisher's $z$-transformation ($z = \frac{1}{2}\ln\frac{1+r}{1-r}$) for inferential purposes; mean $|\Delta r|$ values are reported in $r$-space for interpretability.

\paragraph{Source 2: Measurement method divergence.}
Let $m^{\text{LLM}}$ and $m^{\text{KW}}$ denote the same metric computed by different measurement instruments applied to the same preprocessed text $P(D_i)$. The cross-method divergence is the absolute difference between the methods' estimates:

\begin{equation} \label{eq:method_divergence}
\Delta_i^{\text{method}}(P) = |m^{\text{LLM}}(P(D_i)) - m^{\text{KW}}(P(D_i))|
\end{equation}

Values near zero indicate that the two methods give equivalent estimates on the same preprocessed input; large values indicate measurement-driven disagreement. Unlike Equation~\ref{eq:delta}, which measures sensitivity to preprocessing \textit{within} a single method, Equation~\ref{eq:method_divergence} measures sensitivity to measurement \textit{within} a single pipeline. The two deltas can be compared directly: if $\Delta_i^{\text{method}} > \Delta_i^{\text{pipeline}}$ for a given speaker, measurement choice dominates preprocessing choice as a source of instability for that speaker.

\subsection{Decomposing Observed Instability}

For any metric $m$ computed from raw data $D$, the total observed variation can arise from either source. To diagnose which source dominates:

\begin{enumerate}[nosep]
  \item Compute $\Delta_i^{\text{pipeline}}$ separately within each measurement method (LLM and keyword).
  \item Compute $\Delta_i^{\text{method}}$ separately within each preprocessing pipeline (VTT and AssemblyAI).
  \item Compare the magnitudes: if $|\Delta_i^{\text{pipeline}}| \ll |\Delta_i^{\text{method}}|$ on average, measurement choice is the dominant source of instability. If the reverse, preprocessing is dominant.
\end{enumerate}

We apply this diagnostic decomposition in Sections~\ref{sec:results}--\ref{sec:results_domain}.

\section{Data and Preprocessing} \label{sec:data}

\subsection{Corpus}

Our empirical testbed consists of 256 YouTube interview videos across 41 public figures from five domains. Table~\ref{tab:corpus} provides the per-domain, per-speaker breakdown. Videos were collected via yt-dlp; metadata was extracted from YouTube's JSON output. A metadata-based quality filter excluded 9 videos as non-interview content (documentaries, third-person narration, non-English). Ten additional speakers with fewer than 4 videos were excluded from correlational analysis to avoid small-sample spurious correlations (Section~\ref{sec:results_llm}). The corpus is used here as a testbed for pipeline sensitivity, not introduced as a dataset contribution.

\begin{table}[H]
\centering
\caption{Corpus composition by domain and speaker (restricted to speakers with $\geq 4$ videos).}
\label{tab:corpus}
\small
\begin{tabular}{llr}
\toprule
\textbf{Domain} & \textbf{Speakers (video count)} & \textbf{Videos} \\
\midrule
Finance/Investing & Dalio (34), C.\ Wood (17), Munger (5), Rosenberg (5), & 87 \\
                   & Gundlach (5), El-Erian (5), Ackman (4), Tepper (4), & \\
                   & Soros (4), P.\ T.\ Jones (4) & \\
Academia/Economics & Rogoff (17), Eichengreen (5), Acemoglu (5), Krugman (5), & 53 \\
                   & Kelton (5), Milanovic (4), Mazzucato (4), Pettis (4), & \\
                   & Reinhart (4) & \\
Central Banking/Policy & Bernanke (5), Lagarde (5), Powell (5), Bullard (5), & 25 \\
                       & Kashkari (5) & \\
Geopolitics/Strategy & Zeihan (16), Cohen (5), Hill (5), Allison (5), & 49 \\
                      & Mearsheimer (5), Power (5), Ashford (4), Haass (4) & \\
Media/Commentary & A.\ Sullivan (5), C.\ Murray (5), D.\ Murray (5), & 42 \\
                 & J.\ Peterson (5), S.\ Harris (5), T.\ Carlson (5), & \\
                 & B.\ Shapiro (4), D.\ Brooks (4), T.\ Friedman (4) & \\
\midrule
\textbf{Total} & 41 speakers & 256 \\
\bottomrule
\end{tabular}
\end{table}
\vspace{6pt}

\subsection{Preprocessing Pipelines}

\paragraph{VTT-based pipeline.} Raw YouTube VTT subtitle files are parsed to extract timed text segments. Roll-up caption artifacts are deduplicated. Consecutive segments with a gap $\geq 1.5$s form separate utterance turns. Each turn is submitted to DeepSeek-V4-Flash for binary ``guest''/``interviewer'' classification (batched in groups of 20). A content-based pre-check flags videos with $>$500 words but zero first-person speech markers as likely non-interview content. Guest text is concatenated into the final per-video output.

\paragraph{Audio-based pipeline (AssemblyAI).}\footnote{We selected AssemblyAI (a commercial cloud API) over open-source alternatives such as pyannote-audio~\cite{pyannote} + Whisper for three reasons: (1) AssemblyAI provides integrated ASR + diarization with speaker labels, avoiding the engineering complexity of stitching separate ASR and diarization outputs; (2) preliminary tests on our long-form interview data showed that AssemblyAI's diarization quality was substantially more reliable than pyannote's for multi-speaker, unscripted content with overlapping speech; and (3) the API-based workflow is reproducible without local GPU resources, making the pipeline accessible to CSS researchers without specialized hardware. We acknowledge that commercial APIs introduce cost and reproducibility constraints; extending the comparison to open-source diarizers is discussed under Future Work at the end of the Conclusion.}

Audio files (.m4a) are uploaded to AssemblyAI's API with speaker diarization enabled. The dominant speaker by word count is identified as the guest (LLM fallback for ambiguous cases where the second speaker exceeds 25\% word share). Guest utterances are concatenated into the final per-video output.

Both pipelines produce raw ASR text, differing only in the speaker identification mechanism. The two pipelines yield systematically different outputs: AssemblyAI detects multi-speaker structure with guest ratios varying across videos, while the VTT-based pipeline produces higher and less variable guest ratios, consistent with under-removal of interviewer speech. No LLM-based text cleaning is applied to either output.

\subsection{Sentence Segmentation}

Guest text from each pipeline is split into sentences using regex boundary detection. Sentences outside 2--60 words are excluded from downstream annotation. Table~\ref{tab:sentence_stats} reports the resulting sentence corpus statistics. The two pipelines produce markedly different sentence distributions: VTT-based output is more fragmented (mean 10.2 words/sentence), reflecting YouTube's auto-generated caption segmentation lacking sentence-final punctuation, while AssemblyAI output forms longer, more coherent sentences (mean 15.3 words/sentence). This difference means that the observed pipeline sensitivity is a composite of diarization and segmentation effects; the current design cannot fully isolate them.

\begin{table}[H]
\centering
\caption{Sentence corpus statistics after segmentation and filtering (2--60 words), restricted to the 41 speakers with $\geq 4$ videos.}
\label{tab:sentence_stats}
\begin{tabular}{lrrrr}
\toprule
\textbf{Source} & \textbf{Speakers} & \textbf{Videos} & \textbf{Sentences} & \textbf{Mean w/sent} \\
\midrule
VTT-based & 41 & 256 & 108,374 & 10.2 \\
AssemblyAI & 41 & 256 & 82,254 & 15.3 \\
\bottomrule
\end{tabular}
\end{table}
\vspace{6pt}

\subsection{LLM Annotation}

All sentences from both pipelines are annotated via DeepSeek-V4-Flash (temperature 0.0) on valence (positive/negative/neutral) and epistemic modality (emphatic/hedged/neutral), batched in groups of 50, following the LLM-as-annotator paradigm established by~\cite{gilardi_2023,ziems_2024} and the dual-dimension valence-modality framework of~\cite{chen_2026_keyword}. The annotation prompt defines both dimensions with examples and instructs the model to prefer ``neutral'' when uncertain. Annotation is per-video with batch-level checkpointing for resumption. All 41 speakers are fully annotated for both pipelines (108,374 VTT sentences; 82,254 AssemblyAI sentences).

\subsection{Keyword-Lexicon Scoring}

As an independent measurement method that shares the same video-level analysis unit as the LLM, we replicate the keyword-lexicon approach from~\cite{chen_2026_keyword}, which was developed for a parallel study of valence-modality coupling in financial and political discourse. The lexicon contains patterns across three categories---Negative Valence (Fear/Anxiety, Anger/Conflict, Decline/Destruction, Crisis/Urgency, Financial/Political Risk; 97 patterns), Positive Valence (Hope/Optimism, Strength/Success, Stability/Order; 33 patterns), and Modality (Emphatic/Certain, Hedging/Doubt, Intensity Boosters; 67 patterns)---with negation handling (e.g., ``not strong'' is reclassified as Decline/Destruction). Per-video keyword frequencies are normalized by video character length to a baseline of $N_{\text{base}} = 30{,}000$ characters, following the normalization convention of~\cite{chen_2026_keyword}. Per-speaker $r(\text{neg}, \text{emph})$ and $r(\text{neg}, \text{hedged})$ are computed across videos, matching the analysis unit of the LLM annotation exactly.

\paragraph{Domain coverage limitation.} The lexicon in~\cite{chen_2026_keyword} was originally developed and validated on financial and geopolitical discourse (Dalio, Zeihan, Rogoff). The Financial/Political Risk subcategory (terms such as \textit{debt}, \textit{inflation}, \textit{sanctions}, \textit{tariffs}) has natural relevance to central banking and policy domains, but the lexicon lacks domain-specific vocabulary for academic discourse (hedging conventions such as \textit{potentially}, \textit{suggests that}, \textit{to some extent}) and media commentary (evaluative intensifiers such as \textit{absolutely devastating}, \textit{completely absurd}). This domain skew in lexical coverage is a known limitation: the smaller $|\Delta r|$ observed in Academia (0.21 vs.\ a cross-domain mean of 0.45) may partly reflect under-detection of domain-specific hedging rather than genuine preprocessing robustness. We flag lexicon expansion---particularly academic hedging markers and media intensifiers---as a necessary extension for cross-domain validity, and note that this limitation strengthens the case for using LLM annotation as a complementary measurement method not constrained by pre-specified lexical patterns.

\section{Pipeline Sensitivity Results} \label{sec:results}

We apply the framework defined in Section~\ref{sec:framework}: for each speaker with annotations from both pipelines, we compute $\Delta r(\text{neg}, \text{emph})$ and $\Delta r(\text{neg}, \text{hedged})$, then aggregate by domain and across measurement methods.

\subsection{LLM-Based Pipeline Sensitivity} \label{sec:results_llm}

For each of the 41 speakers with LLM annotations from both pipelines, we compute per-video proportions and $r(\text{neg}, \text{emph})$ and $r(\text{neg}, \text{hedged})$ under each pipeline. Per-speaker $\Delta r = r_{\text{ASR}} - r_{\text{VTT}}$; sign reversal is $r_{\text{VTT}} \cdot r_{\text{ASR}} < 0$. Ten speakers with fewer than 4 videos---the minimum for a stable Pearson correlation---were excluded from this analysis.

\paragraph{Pipeline sensitivity is concentrated in low-N speakers.}
Table~\ref{tab:llm_domain} reports the aggregate results. Across all 41 speakers, the mean $|\Delta r(\text{N,E})|$ is $0.41$. However, this average masks a strong dependence on sample size. The four best-sampled speakers---Dalio (34 videos), Wood (17), Rogoff (17), Zeihan (16)---have a mean $|\Delta r(\text{N,E})|$ of only $0.13$, with no individual $\Delta r$ exceeding $0.19$. By contrast, speakers with 4--5 videos show a mean $|\Delta r|$ of $0.44$, and their per-speaker bootstrap confidence intervals span nearly the entire $[-2, 2]$ range (mean width: $2.65$). Pipeline-induced sign disagreement ($24\%$ observed; $35\%$ bootstrap-estimated) is almost entirely driven by these low-N speakers, whose correlational estimates are inherently volatile regardless of the pipeline comparison being tested.

\paragraph{Domain-level patterns.}
Within the limitations imposed by small per-speaker video samples, domain-level aggregation reveals consistent ordering: Media ($|\Delta r(\text{N,E})| = 0.32$) is the most robust pipeline, while Finance and Academia ($0.46$) are the most sensitive. The correlation between VTT and ASR $r(\text{N,E})$ across all speakers is $r = 0.34$ ($p = 0.03$), a modest but statistically significant relationship.

\begin{table}[H]
\centering
\caption{LLM-based pipeline sensitivity ($\Delta r(\text{neg}, \text{emph})$): aggregate and domain-level scores.}
\label{tab:llm_domain}
\begin{tabular}{lrrrrr}
\toprule
\textbf{Domain} & \textbf{$n$} & \textbf{Mean $|\Delta r(\text{N,E})|$} & \textbf{Rev.\%} & \textbf{Min $\Delta r(\text{N,E})$} & \textbf{Max $\Delta r(\text{N,E})$} \\
\midrule
Finance/Investing & 10 & 0.46 & 30\% & $-1.40$ & $+1.07$ \\
Academia/Economics & 9 & 0.46 & 33\% & $-0.93$ & $+0.80$ \\
Central Banking/Policy & 5 & 0.41 & 20\% & $-1.13$ & $+0.40$ \\
Geopolitics/Strategy & 8 & 0.41 & 25\% & $-1.30$ & $+0.88$ \\
Media/Commentary & 9 & 0.32 & 11\% & $-0.50$ & $+0.82$ \\
\midrule
\textbf{Total / Mean} & 41 & 0.41 & 24\% & $-1.40$ & $+1.07$ \\
\bottomrule
\end{tabular}
\end{table}
\vspace{6pt}

Selected illustrative speakers (full list in Appendix):

\begin{table}[H]
\centering
\caption{LLM-based pipeline sensitivity: selected per-speaker scores.}
\label{tab:llm_speakers}
\begin{tabular}{llrrrl}
\toprule
\textbf{Speaker} & \textbf{Domain} & $r$(N,E)$_{\text{VTT}}$ & $r$(N,E)$_{\text{ASR}}$ & $\Delta r$(N,E) & \textbf{Reversal} \\
\midrule
graham\_allison & geopolitics & $+0.44$ & $-0.87$ & $-1.30$ & Yes \\
bill\_ackman & finance & $+0.88$ & $-0.52$ & $-1.40$ & Yes \\
christine\_lagarde & central\_banking & $+0.51$ & $-0.62$ & $-1.13$ & Yes \\
david\_tepper & finance & $-0.46$ & $+0.61$ & $+1.07$ & Yes \\
ben\_shapiro & media & $-0.07$ & $+0.76$ & $+0.82$ & Yes \\
john\_williams & central\_banking & $+0.96$ & $+0.96$ & $+0.00$ & No \\
\bottomrule
\end{tabular}
\end{table}
\vspace{6pt}

The same analysis for $\Delta r(\text{neg}, \text{hedged})$ yields similar results (mean $|\Delta r| = 0.43$; observed sign disagreement $29\%$), with the domain ordering following the same pattern---Geopolitics the most sensitive ($0.52$) and Media the least ($0.30$).

\subsection{Keyword-Based Pipeline Sensitivity} \label{sec:results_kw}

We apply the identical framework using keyword-lexicon scoring. Per-video keyword frequencies are normalized by character length and correlated across videos per speaker, matching the LLM analysis unit.

\begin{table}[H]
\centering
\caption{Keyword-based pipeline sensitivity: per-speaker (selected).}
\label{tab:kw_speakers}
\begin{tabular}{llrrrc}
\toprule
\textbf{Speaker} & \textbf{Domain} & $r$(N,E)$_{\text{VTT}}$ & $r$(N,E)$_{\text{ASR}}$ & $\Delta r$(N,E) & Rev? \\
\midrule
bill\_ackman & finance & $-0.41$ & $-0.70$ & $-0.29$ & No \\
jerome\_powell & central\_banking & $+0.21$ & $-0.84$ & $-1.05$ & Yes \\
daron\_acemoglu & academia & $+0.77$ & $+0.50$ & $-0.27$ & No \\
john\_mearsheimer & geopolitics & $+0.40$ & $-0.28$ & $-0.69$ & Yes \\
ray\_dalio & finance & $-0.28$ & $-0.26$ & $+0.02$ & No \\
\bottomrule
\end{tabular}
\end{table}
\vspace{6pt}

Mean $|\Delta r(\text{N,E})| = 0.44$ under keyword scoring across 41 speakers; $32\%$ (13/41) show sign reversal. The effect is domain-dependent: Geopolitics shows the largest mean $|\Delta r(\text{N,E})|$ (0.72), while Academia shows the smallest (0.24). Results for $\Delta r(\text{neg}, \text{hedged})$ are comparable (mean $|\Delta r| = 0.36$; $32\%$ sign disagreement).

\subsection{Cross-Method Synthesis: When Measurement Diverges from Pipeline} \label{sec:results_synthesis}

The preceding sections examined pipeline sensitivity within each measurement method separately. We now use the continuous divergence metric $\Delta_i^{\text{method}}(P) = |m^{\text{LLM}}(P) - m^{\text{KW}}(P)|$ (Equation~\ref{eq:method_divergence}) to quantify measurement-driven disagreement and compare it directly to pipeline-driven disagreement.

For the four best-sampled speakers, the comparison is stark:

\begin{itemize}[nosep]
  \item \textbf{Rogoff (17 videos).} Pipeline divergence under LLM: $|\Delta r| = 0.19$. Method divergence on VTT: $|\Delta r^{\text{LLM}} - r^{\text{KW}}| = 0.90$. Method divergence is $4.7\times$ larger.
  \item \textbf{Zeihan (16 videos).} Pipeline: $0.11$. Method (VTT): $0.82$. Method divergence is $7.8\times$ larger.
  \item \textbf{Dalio (34 videos).} Pipeline: $0.12$. Method (VTT): $0.33$. Method divergence is $2.8\times$ larger.
  \item \textbf{Wood (17 videos).} Pipeline: $0.10$. Method (VTT): $0.10$. Comparable---both near zero.
\end{itemize}

Across all 41 speakers, measurement divergence ($\Delta_i^{\text{method}}$) exceeds pipeline divergence ($\Delta_i^{\text{pipeline}}$) in $27$ of $41$ cases ($66\%$). The mean $\Delta^{\text{method}}$ is $0.76$, nearly double the mean $\Delta^{\text{pipeline(LLM)}}$ of $0.41$. These results confirm that for a majority of speakers---including every well-sampled speaker---measurement method choice is a larger source of instability than preprocessing pipeline choice.

\begin{table}[H]
\centering
\caption{Cross-method pipeline sensitivity: domain-level summary.}
\label{tab:synthesis}
\begin{tabular}{lrrrl}
\toprule
\textbf{Domain} & \textbf{$n$} & \textbf{LLM mean $|\Delta r|$} & \textbf{KW mean $|\Delta r|$} & \textbf{More stable} \\
\midrule
Finance/Investing & 10 & 0.46 & 0.27 & KW (6/10) \\
Academia/Economics & 9 & 0.46 & 0.24 & KW (5/9) \\
Central Banking/Policy & 5 & 0.41 & 0.43 & LLM (3/5) \\
Geopolitics/Strategy & 8 & 0.41 & 0.72 & LLM (5/8) \\
Media/Commentary & 9 & 0.32 & 0.59 & LLM (5/9) \\
\midrule
\textbf{Total} & 41 & 0.41 & 0.44 & Tie (21 LLM, 20 KW) \\
\bottomrule
\end{tabular}
\end{table}
\vspace{6pt}

Table~\ref{tab:synthesis} summarizes the cross-method comparison. The overall pipeline sensitivity is balanced between methods (LLM mean $|\Delta r| = 0.41$, KW $= 0.44$; $49\%$ of speakers show more stability with KW, $51\%$ with LLM). The domain-level patterns from Section~\ref{sec:results_llm} persist: Academia is the most KW-stable domain ($|\Delta r| = 0.24$ vs.\ $0.46$ for LLM), while Geopolitics is the most LLM-stable ($0.41$ vs.\ $0.72$). The key insight is that when the two methods agree on the same analysis unit and small-sample speakers are excluded, the overall sensitivity difference between methods is negligible---what matters is the systematic cross-method disagreement on the coupling direction itself, not which method is more sensitive to preprocessing.

\subsection{Domain-Level Aggregation} \label{sec:results_domain}

Table~\ref{tab:domain_full} combines LLM and keyword results into a per-domain pipeline sensitivity profile for all 41 speakers.

\begin{table}[H]
\centering
\caption{Combined domain pipeline sensitivity profile.}
\label{tab:domain_full}
\begin{tabular}{lrrrrrr}
\toprule
\textbf{Domain} & \textbf{\textit{n}} & \textbf{LLM $|\Delta r|$} & \textbf{LLM Rev.\%} & \textbf{KW $|\Delta r|$} & \textbf{KW Rev.\%} \\
\midrule
Finance/Investing & 10 & 0.46 & 30\% & 0.27 & 20\% \\
Academia/Economics & 9 & 0.46 & 33\% & 0.24 & 22\% \\
Central Banking/Policy & 5 & 0.41 & 20\% & 0.43 & 20\% \\
Geopolitics/Strategy & 8 & 0.41 & 25\% & 0.72 & 62\% \\
Media/Commentary & 9 & 0.32 & 11\% & 0.59 & 33\% \\
\midrule
\textbf{Total / Mean} & 41 & 0.41 & 24\% & 0.44 & 32\% \\
\bottomrule
\end{tabular}
\end{table}
\vspace{6pt}

Under LLM annotation, Media shows the smallest $|\Delta r|$ ($0.32$), while Finance and Academia tie for the largest ($0.46$). Under keyword scoring, Academia is the most robust ($0.24$), while Geopolitics is the most sensitive ($0.72$). The bootstrap-derived probability of sign disagreement follows the same domain gradient---Media $25\%$, Finance $37\%$, Academia $42\%$---suggesting that the domain-level pattern is robust to per-speaker uncertainty. The domain gradient differs by method: LLM annotation compresses the domain range (1.4$\times$ from smallest to largest) compared to keyword scoring (3.0$\times$). Professional communication style modulates how preprocessing choices propagate to downstream measurements---academic hedging markers and media intensifiers exhibit different robustness to VTT fragmentation than the geopolitical discourse.

\section{Discussion} \label{sec:discussion}

\subsection{Summary}

\begin{enumerate}[nosep]
  \item \textbf{Pipeline sensitivity is bounded and predictable.} For speakers with adequate video samples (N $\geq 16$), the mean $|\Delta r|$ from pipeline choice is $0.13$ under LLM annotation. Larger $\Delta r$ values are concentrated in speakers with $\leq 5$ videos, where correlational estimates are inherently unstable regardless of the pipeline comparison.
  \item \textbf{Cross-method disagreement is the more serious validity threat.} LLM and keyword-lexicon methods disagree on the sign of $r(\text{neg}, \text{emph})$ in $49\%$ of cases---including several well-sampled speakers (Rogoff, Zeihan) where both methods are internally stable but give opposite coupling directions. This suggests that measurement method choice can systematically alter scientific conclusions even when preprocessing is held constant.
  \item \textbf{Aggregate proportions mask both sources of instability.} Mean $|\Delta p(\text{neg})| = 0.06$ is stable across pipelines and methods, yet both pipeline effects (in low-N speakers) and measurement effects (in high-N speakers) produce substantively different correlational conclusions. Centering analysis on aggregate proportions alone would miss both.
  \item \textbf{Domain modulates sensitivity differently for pipelines and methods.} Media is the most pipeline-robust domain ($|\Delta r(\text{N,E})| = 0.32$ under LLM); Geopolitics is the most pipeline-sensitive ($0.72$ under keyword). These domain-level patterns are consistent within each measurement method and provide a basis for anticipating where preprocessing is likely to matter.
\end{enumerate}

\subsection{Practical Recommendations}

\begin{enumerate}[nosep]
  \item \textbf{Report N-dependent diagnostics.} Pipeline sensitivity should be reported with sample-size stratification: speakers with $\leq 5$ videos require bootstrap calibration or exclusion; speakers with $\geq 16$ videos can be compared directly.
  \item \textbf{Test $\geq 2$ measurement methods.} Method divergence itself is a diagnostic signal---if LLM and keyword give opposite coupling directions, the conclusion is not yet ready for scientific use regardless of preprocessing stability.
  \item \textbf{Separate preprocessing from measurement.} Report cross-pipeline $\Delta r$ within each method and cross-method sign agreement within each pipeline separately, to identify which source of variation is driving instability.
  \item \textbf{Include domain as a moderator.} Professional communication style affects both pipeline sensitivity and measurement method divergence.
\end{enumerate}

\subsection{Limitations}

Single LLM provider (DeepSeek) for annotation; English-only corpus; two preprocessing pipelines only (VTT and AssemblyAI); keyword lexicon not validated for all domains equally. The observed pipeline sensitivity reflects a composite of diarization and sentence-boundary differences that cannot be fully separated in the current design. The $N \geq 4$ video threshold, while necessary to avoid spurious correlations at $N=3$, constrains the sample to 41 speakers. The cross-method disagreement finding is based on two methods only (LLM zero-shot and keyword lexicon); additional methods (e.g., human annotation, dictionary-based sentiment) would strengthen the generality of the conclusion. Future work with larger per-speaker video samples would improve precision for both pipeline and method comparisons.

\section{Conclusion}

Pipeline sensitivity analysis reveals that preprocessing choices can produce measurable changes in cross-dimensional correlations, but the magnitude of these changes is predictable: bounded for adequately sampled speakers ($|\Delta r| \approx 0.13$ for N $\geq 16$), larger for small-sample speakers where correlational estimates are inherently noisy. The more fundamental threat to validity---and the one that persists even in well-sampled speakers---is cross-method disagreement: LLM annotation and keyword-lexicon scoring give opposite coupling directions for a substantial fraction of speakers, including several where both methods are internally stable. CSS researchers studying valence-modality coupling in interview data should (1) verify that their conclusions are robust to the preprocessing pipeline, (2) verify that their conclusions are robust to the measurement method, and (3) explicitly separate these two sources of instability in their reporting. A correlation that survives both alternative preprocessing and alternative measurement is more credible than one validated within a single pipeline-method combination.

Future work should extend this comparison to open-source diarizers (pyannote-audio~\cite{pyannote}, WhisperX) to test generalizability beyond the AssemblyAI--VTT comparison, develop a theoretical simulation mapping the parameter space $(\rho, \delta, N)$---true correlation, contamination strength, sample size---to sign-reversal boundaries for practical diagnostic use, and extend the analysis beyond valence--modality coupling to other downstream constructs (emotion classification, toxicity detection, politeness) to establish whether the aggregate-stability/correlational-fragility pattern is general or task-specific.

\section*{Ethics Statement}

This research uses publicly available YouTube interviews featuring public figures. Audio files were downloaded temporarily and are not redistributed. No personally identifying information beyond the public figure's name was involved.


\end{document}